\documentclass[letterpaper]{article} 
\usepackage[]{anonymous}  
\usepackage{times}  
\usepackage{helvet}  
\usepackage{courier}  
\usepackage[hyphens]{url}  
\usepackage{graphicx} 
\usepackage{natbib}  
\usepackage{caption} 
\usepackage{algorithm}
\usepackage{algorithmic}
\usepackage{amsmath} 
\usepackage{newfloat}
\usepackage{listings}
\usepackage{nameref}
\usepackage{multirow}
\DeclareCaptionStyle{ruled}{labelfont=normalfont,labelsep=colon,strut=off} 
\floatstyle{ruled}
\newfloat{listing}{tb}{lst}{}
\floatname{listing}{Listing}
\title{StyleSentinel: Reliable Artistic Copyright Verification via Stylistic Fingerprints}
\author {
	Lingxiao Chen\textsuperscript{\rm 1},
        Liqin Wang\textsuperscript{\rm 1},
	Wei Lu\textsuperscript{\rm 1}\thanks{Corresponding authors}
}
\affiliations {
	\textsuperscript{\rm 1}School of Computer Science and Engineering, Ministry of Education Key Laboratory of Information Technology, Guangdong Province Key Laboratory of Information Security Technology, Sun Yat-sen University, Guangzhou 510006, China\\
	chenlx67@mail2.sysu.edu.cn, wanglq37@mail2.sysu.edu.cn, luwei3@mail.sysu.edu.cn
}

\usepackage{bibentry}

\begin{document}

\maketitle

\begin{abstract}

The versatility of diffusion models in generating customized images has led to unauthorized usage of personal artwork, which poses a significant threat to the intellectual property of artists. Existing approaches relying on embedding additional information, such as perturbations, watermarks, and backdoors, suffer from limited defensive capabilities and fail to protect artwork published online. 
In this paper, we propose StyleSentinel, an approach for copyright protection of artwork by verifying an inherent stylistic fingerprint in the artist's artwork. 
Specifically, we employ a semantic self-reconstruction process to enhance stylistic expressiveness within the artwork, which establishes a dense and style-consistent manifold foundation for feature learning. 
Subsequently, we adaptively fuse multi-layer image features to encode abstract artistic style into a compact stylistic fingerprint.
Finally, we model the target artist's style as a minimal enclosing hypersphere boundary in the feature space, transforming complex copyright verification into a robust one-class learning task.
Extensive experiments demonstrate that compared with the state-of-the-art, StyleSentinel achieves superior performance on the one-sample verification task.
We also demonstrate the effectiveness through online platforms.

\end{abstract}


\section{Introduction}

The versatility of diffusion models such as DALL-E \cite{ramesh2021zero}, Midjourney, Kandinsky \cite{razzhigaev2023kandinsky}, and Stable Diffusion \cite{rombach2022high} represents a revolutionary advance in generative AI. They can transform text into detailed and stylistically diverse images. However, simple prompts fail to meet specific requirements and full model retraining remains prohibitively expensive. Fine-tuning techniques like LoRA \cite{hu2021loralowrankadaptationlarge}, Dreambooth \cite{ruiz2023dreambooth}, and Textual Inversion \cite{gal2022image} democratize AI art by enabling large foundation models to learn specific concepts, objects, or styles from small image sets, significantly lowering technical barriers.
\begin{figure}[t]
\includegraphics[width=\linewidth]{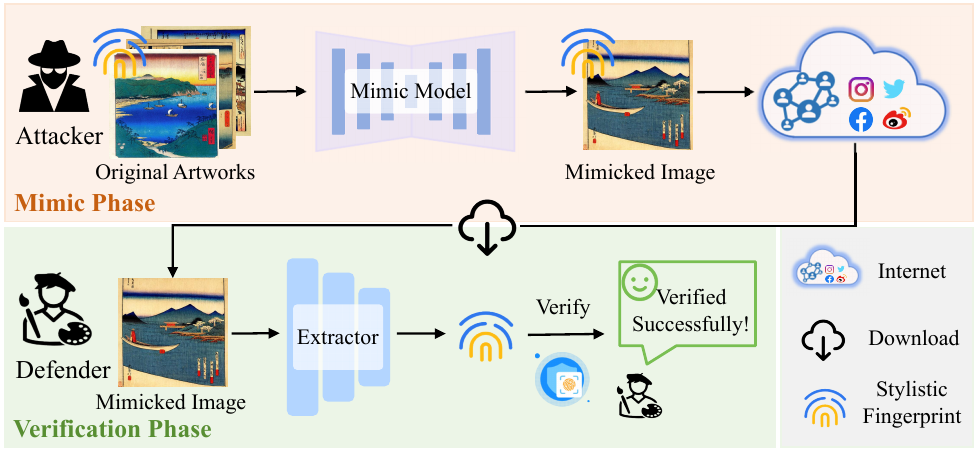}
\caption{Illustration of our approach for copyright verification. The stylistic fingerprint in the artwork is preserved during the mimic phase, and it can be extracted as robust evidence for copyright verification.}
\label{intro}
\end{figure}

This powerful capability for visual understanding and style transfer allows users to create personalized artworks. However, it also creates a convenient way for unauthorized usage of artworks. Malicious attackers can easily scrape unauthorized artworks from public sources and rapidly fine-tune a model on consumer-grade GPUs to mimic the styles. It constitutes plagiarism of the artwork, devaluing the original art while directly threatening intellectual property rights and economic interests \cite{moayeri2024rethinking}.

Current research on artwork copyright protection can be primarily categorized into three types of methods. Perturbation-based methods aim to disrupt the output of malicious models \cite{chen2024editshield,shan2023glaze,van2023anti,zhao2024can}, typically achieved by injecting imperceptible perturbations into the images. These perturbations are designed to corrupt the training process, causing distorted and stylistically incongruous output. Watermark-based methods \cite{cui2023diffusionshield,luo2023steal,ma2023generative,zhu2023detection} embed invisible watermarks into images. Models trained on such images will produce outputs with these watermarks, enabling copyright verification by detecting them. Backdoor-based methods \cite{wang2023diagnosis,chou2023backdoor} introduce backdoors into datasets, which allow infringement detection by testing whether a model exhibits specific backdoor behaviors triggered by pre-defined inputs.

However, both perturbation-based, watermark-based, and backdoor-based methods are subject to critical limitations. First, they rely on embedding extrinsic signals into images, but the instability of the signals results in limited defensive capabilities against potential attacks.
Second, these pre-processing dependent methods cannot retroactively protect artworks already circulating online without prior safeguards.

To address these, as illustrated in Figure \ref{intro}, we propose StyleSentinel to verify the copyright by extracting a shared stylistic fingerprint from an artist's artworks. Unlike existing approaches that embed invisible signals into images, our solution neither compromises visual quality nor requires preprocessing. Our work is motivated by two key observations. First, the images generated by the diffusion models exhibit significant visual and stylistic similarities with the training data \cite{wang2023alteration,somepalli2023diffusion}, suggesting that style functions as a persistent fingerprint preserved during training. Second, the distinct styles possessed by most artists differentiate their artworks from others \cite{moayeri2024rethinking}. This inherent distinctiveness makes the extraction of such stylistic fingerprints feasible.

Specifically, considering the limited number of artworks available for a single artist, we employ a semantics self-reconstruction process to enhance stylistic expressiveness within the artworks. This process leverages semantics and style of an image to synthesize new augmented samples, which establishes a style-consistent and data-rich manifold foundation for subsequent feature learning. 
Subsequently, due to the multi-level nature of artistic style, we design a multi-layer attention style extractor to generate a compact stylistic fingerprint from each image. This module extracts features from multiple layers of a backbone network and adaptively fuses them into a style feature vector, encoding a discriminative fingerprint for subsequent verification.
Finally, we use a hypersphere-based verifier to model stylistic boundaries, enabling reliable verification of the extracted stylistic fingerprint. This component learns a minimal enclosing boundary for all style embeddings of the target artist, transforming copyright verification into a more robust one-class learning task.
During verification, copyright attribution is determined simply by checking whether a suspect image's style vector falls within the prelearned hypersphere boundary. In summary, our contributions are as follows:
\begin{itemize}
    \item We propose StyleSentinel for artistic copyright protection by verifying inherent artistic fingerprints from artworks. It requires no image preprocessing and enables verification for unprotected images published online.
    \item We introduce a semantic self-reconstruction process to strengthen the data foundation for stylistic feature learning. We further design a multi-layer attention style extractor with a hypersphere-based verifier to encode stylistic fingerprint and achieve reliable verification.
    \item Extensive experiments demonstrate that StyleSentinel achieves superior performance against state-of-the-art baselines in the one-sample verification task. The effectiveness is further validated on two real-world platforms.
\end{itemize}


\section{Related Works}
\subsection{Style Mimicry with Diffusion Models}
Large-scale pre-trained diffusion models, trained on massive image-text pair datasets like LAION \cite{NEURIPS2022_a1859deb}, demonstrate remarkable capabilities in generating high-quality and diverse images. However, the prohibitive cost of retraining these foundation models has limited the proliferation of their use for stylistic mimicry.

To enable low-cost personalized generation, the research community has developed efficient fine-tuning techniques, which have led to the rampant problem of style mimicry. Among these, DreamBooth and LoRA deliver the most potent style replication. DreamBooth, a representative high-fidelity method, achieves sophisticated style learning with minimal images (typically 3-5) by exposing the model to a unique identifier (e.g., a rare token [V]) and fine-tuning the entire U-Net. It further mitigates language drift and overfitting through class-specific prior preservation loss. In contrast to DreamBooth’s extensive weight adjustments, LoRA freezes all pretrained parameters and operates via parallel low-rank injections into key modules. With trainable parameters representing merely 0.01\%-0.1\% of the original model, LoRA dramatically reduces computational and storage demands. While the lightweight and efficient nature of these techniques has profoundly democratized AI art, it paradoxically enables unauthorized and low-cost artistic style replication, posing significant threats to creative rights.

\subsection{Dataset Copyright Protection}
Current dataset copyright protection for generative models relies on embedding external information. Perturbation-based methods introduce adversarial noise, impair the semantic learning of the model and cause corrupted outputs. However, they are vulnerable to defenses like adversarial purification \cite{NEURIPS2023_222dda29}, where preprocessing can nullify the effect. Watermark-based methods embed covert signals designed to persist through training and appear in outputs, while backdoor-based methods implant triggers for detection. Crucially, both watermark-based and backdoor-based methods face the challenge of ensuring signal robustness against image transformations and the training process \cite{NEURIPS2024_10272bfd} while preserving host image quality.

Some approaches explore a distinct path requiring no image modification. However, approaches like membership inference \cite{chen2020gan,shokri2017membership} are often impractical, as they require massive sample generation for statistical analysis. ArtistAuditor \cite{du2025artistauditor} addresses to leverage style representation for copyright verification, but it fails to perform adequately in one-sample verification scenarios.

\begin{figure*}
\centering
{\includegraphics[width=0.95\linewidth]{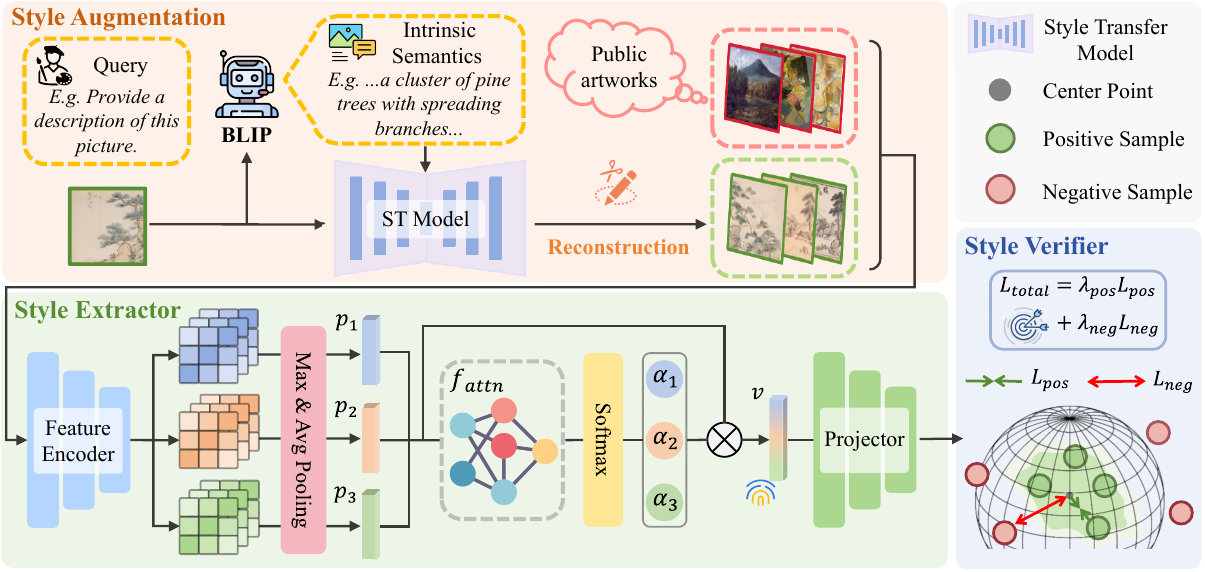}}
\caption{Overview of the proposed StyleSentinel. The pipeline comprises three main stages. (1) Style Augmentation: A semantic self-reconstruction process generates style-consistent augmented data from the original works. (2) Style Extraction: A multi-layer attention style extractor encodes each artwork into a compact style fingerprint. (3) Style Verifier: A hypersphere-based verifier performs one-class classification with the style fingerprint.}
\label{overview}
\end{figure*}

\section{Problem Statement}
\subsection{Threat Model}
\textbf{Attacker’s Goal and Capability. }
Attackers aim to generate new images that closely mimic or replicate the artist's distinctive style for low-cost appropriation and commercial exploitation. Attackers can easily aggregate an artist's publicly available artworks from online portfolios, galleries, and social media platforms to construct a dataset. They potentially preprocess the images with methods like secondary fine-tuning, prompt attacks, or data augmentation. With this dataset, attackers can fine-tune mimic models on consumer-grade GPUs. In most cases, attackers solely publish the generated images while deliberately concealing the mimic model and its implementation details.

\textbf{Defender’s Goal and Capability. }
The defender's primary objective is to reliably verify the copyright of a suspect artwork based on minimal evidence, ideally a single instance. The verification is performed under a reasonable and challenging assumption: The defender acquires the suspect image accidentally, without any additional information about the mimic model or the text prompt used.

\subsection{Design Challenges} \label{Design Challenges}
To achieve reliable verification with stylistic fingerprints for copyright protection, we face two main challenges.
The first challenge is to accurately represent artistic style. An abstract artistic style is generally composed of various elements, and each artist has their own focus.
For example, Claude Monet focuses on the overall atmosphere created by light and shadow, whereas Vincent van Gogh emphasizes local expression through distinctive brushwork and color. Our method must extract a robust stylistic fingerprint from these hierarchical distinctions.

The second challenge is the inherent data sparsity. An individual artist's portfolio is typically a limited corpus, often comprising several tens to hundreds of works. Consequently, our model must learn a generalizable stylistic signature from the limited dataset, avoiding overfitting to the specific content of the training images.

\section{Methods}
\subsection{Overview}
We aim to extract a robust stylistic fingerprint for reliable verification from the artworks.
As illustrated in Figure \ref{overview}, our pipeline comprises three sequential stages. Initially, we introduce a novel module of style augmentation. This module employs a semantic self-reconstruction process to synthesize augmented samples that are not only highly consistent in style with the original artwork but also rich in detailed variations. Subsequently, a multi-layer attention style extractor is used to extract hierarchical features from the images and adaptively fuse them into a single feature vector that holistically represents the style of the image. Finally, the extracted style vector is passed to a hypersphere-based style verifier, which performs verification by determining whether the style features of an unknown sample fall within the decision boundary defined by a hypersphere.

\subsection{Semantic Self-Reconstruction Style Augmentation}
Recalling design challenges, to address the inherent scarcity of data, an effective approach is data augmentation. However, traditional methods such as rotation, compression, and color jittering are insufficient because they operate at a superficial geometric or pixel level randomly. They introduce limited variations and probably disrupt the model's internal representation of artistic style.

To introduce diverse augmented samples without compromising the integrity of the original artistic style, we propose a novel augmentation approach motivated by the insight that artistic style is intrinsically linked to its semantics. An artist's style emerges from their sustained depiction of specific subjects, compositional choices, and atmospheric rendering. For example, Claude Monet's impressionist style of rapid and loose brushstrokes is a result of his semantic focus on capturing the fleeting effects of light on landscapes and water. Preserving this inherent style-semantic coupling is essential for creating meaningful augmentations. Based on this insight, we defined our augmentation approach as the artistic reinterpretation of an image's intrinsic semantics using its own stylistic representation. This can generate high-quality augmented samples that preserve strict style consistency while varying contextual details. Specifically, for an image $X_{ori}$ to be protected, we perform the following two steps to achieve data augmentation.

\begin{figure}[t]
\includegraphics[width=\linewidth]{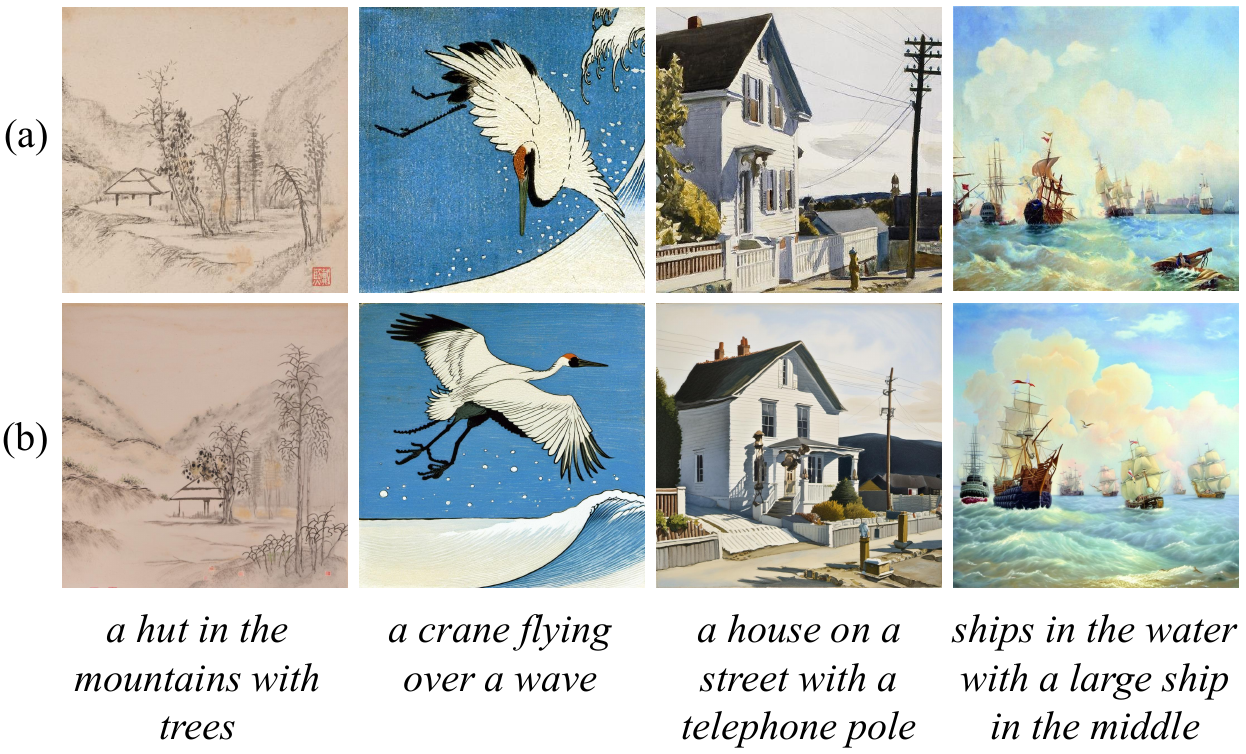}
\caption{Some original images (a) and reconstructed images (b) with their intrinsic
semantics.}
\label{method}
\end{figure}

\textbf{Intrinsic Semantic Extraction. }Prior work \cite{wang2024instantstyle} has demonstrated that, in contrast to the ambiguous nature of stylistic attributes, semantics can often be effectively represented by natural language text. Building upon this insight, we propose using the textual description of an image as its intrinsic semantics. To achieve this, we use BLIP model \cite{li2022blip} to generate a textual description \(T_{ori}\) of the image \(X_{ori}\) serving as its intrinsic semantic information.

\textbf{Self-Reconstruction. }
With intrinsic semantics, we then perform a reconstruction process to generate augmented images. Specifically, we use \(X_{ori}\) as the style reference and \(T_{ori}\) as the content guide, both of which are fed into a style transfer model \cite{wang2024instantstyle} to synthesize the augmented images.
This step leverages the inherent stochasticity of the generative model to introduce valuable generative variance, while strictly preserving the style-semantic coupling (Figure \ref{method}). Consequently, we create augmented images that are both stylistically faithful and diverse. They establish a dense and style-consistent manifold foundation, improving the robustness of subsequent feature learning.


\subsection{Multi-Layer Attention Style Extractor}
As analyzed in design challenges, artistic style is abstract and difficult to quantify.
It constitutes a complex and cross-layer visual pattern, manifesting not only in low-level attributes such as textures and brushstrokes but also in mid-level compositional arrangements and object deformations, as well as in the high-level atmosphere. 
To learn a comprehensive style representation, we design a multi-layer attention style extractor to capture image features across multiple hierarchical levels, and fuse them into a final feature vector that accurately reflects the style.
The style extractor is constructed of two fundamental blocks, a multi-layer feature extraction backbone and an attentional fusion module.

\textbf{Multi-layer Feature Extraction Backbone. }Inspired by \cite{zhang2022domain}, we fine-tune the VGG-19 \cite{simonyan2014very} architecture, which is pre-trained on ImageNet \cite{deng2009imagenet}. We select the output feature maps from three specific layers corresponding to low, mid, and high level features. For each selected feature map, we apply parallel average and max pooling operations to extract both global average and peak feature activations. The outputs of these operations are then concatenated and processed by a final convolutional layer to produce the level-specific feature encoding.

\textbf{Attentional Fusion Module. }After obtaining the feature encodings from different hierarchical levels, an effective approach is to fuse them into a single final style representation. Simple averaging or concatenation operations are suboptimal because they assign equal weight to all feature levels, ignoring the reality that their relative importance varies significantly with the specific artistic style. For instance, a style characterized by prominent brushstrokes might rely more heavily on low-level features, whereas a style defined by a unique atmosphere could be more correlated with high-level features \cite{gatys2015neural}. To this end, we design an attentional fusion module that dynamically learns the weights for different hierarchical feature encodings based on the input. 

Specifically, given the \(N\) feature encodings $\boldsymbol{c}_{1},\ldots,\boldsymbol{c}_{N}$ each with a potentially different dimension, we first unify their dimensionality by projecting each $\boldsymbol{c}_{i}$ into a common $m$-dimensional space using a linear layer, yielding a set of projected vectors $\boldsymbol{p}_{i}$. These vectors are concatenated to form a single tensor $\boldsymbol{P}\in R^{N \times m}$ and passed through a multi-layer perceptron $f_{attn}$ to generate attention scores. Then, these scores are normalized via Softmax function to compute the final attention weights $\boldsymbol{\alpha}=[\alpha_{1},\alpha_{2},\ldots,\alpha_{N}]$ as:
\begin{equation}
\boldsymbol{\alpha} = \text{softmax}(f_{\text{attn}}(\boldsymbol{P}))
\end{equation}
The final feature vector $\boldsymbol{v}$, as the stylistic fingerprint, is calculated as the weighted sum of the projected feature vectors, where each $\boldsymbol{p}_{i}$ is modulated by its corresponding attention weight $\alpha_{i}$:
\begin{equation}
\boldsymbol{v}=\sum_{i=1}^N \alpha_{i}\boldsymbol{p}_{i}
\end{equation}

\subsection{Hypersphere-based Style Verifier}
By far, we have obtained a feature vector $\boldsymbol{v}$ to represent the style. The subsequent step is to verify if this fingerprint belongs to the protected artist. A conventional approach is to construct a binary classifier and train it with a binary cross-entropy loss to learn a separating hyperplane. However, in our specific scenario, the negative class (i.e., non-target artist styles) is inherently diverse and unbounded. It encompasses an infinite spectrum ranging from the styles of all other artists to photographic works and even random images. It is challenging to define a distinct separating boundary for a class that lacks a fixed distribution. This could compel the binary classifier to learn an excessively complex or poorly generalizing decision boundary.

Therefore, we learn a compact description of the positive class, transforming the conventional binary classification problem into a more suitable one-class problem \cite{ruff2018deep}. Specifically, we project the feature vectors obtained in the previous step into a feature space where positive samples are clustered around a center point, forming a minimal-volume hypersphere. In contrast, all negative samples are repelled to the exterior of this hypersphere. Given $N$ samples $\boldsymbol{v}_{1}, \ldots ,\boldsymbol{v}_{N}$ of feature vectors, the loss function for positive samples is defined as:
\begin{equation}
L_{pos}=\frac{1}{N}\sum_{i=1}^N ( \sqrt{d_{i}^{2}+1}-m)
\end{equation}
where $d_{i}=\Vert \phi(\boldsymbol{v}_{i})-o \Vert_2$ is the Euclidean distance, $m$ defines a soft margin, $\phi$ represents the projection module and $o$ is the learnable center point of the hypersphere. We aim to minimize the distance from the style feature vectors to the center $o$ in the projected space. For negative samples, the loss is formulated as:
\begin{equation}
L_{neg}=-\frac{1}{N}\sum_{i=1}^N \log ( 1-\exp( -\beta \cdot ( \sqrt{d_{i}^{2}+1}-m)) +\epsilon)
\end{equation}
where $\beta$ is a hyperparameter that governs the repulsion intensity, and $\epsilon$ is a constant for the numerical stability. The goal is to penalize style vectors that are insufficiently distant from the center, effectively repelling them. The total training objective is defined as:
\begin{equation}
L_{total}=\lambda_{pos}L_{pos}+\lambda_{neg}L_{neg}
\end{equation}
where $\lambda_{pos}$ and $\lambda_{neg}$ are corresponding hyperparameters. The minimization of overall compels the model to sculpt an optimal feature space. The space is organized so that positive samples are enclosed within the hypersphere, while negative samples are expelled to the outside, achieving a fine-grained characterization and separation of the target artistic style.

\subsection{Training and Inference}
Regarding the preparation of the training dataset, for the negative samples, we recommend selecting from large-scale public artwork datasets. The complex and diverse artistic styles contained within such a dataset benefit the construction of a well-defined decision boundary. For the positive samples, we select the artworks of the target artist. If any of these works are present in the negative dataset, they are excluded from it. We then augment the positive samples by style augmentation based on intrinsic semantics, in conjunction with conventional techniques such as random flipping, JPEG compression, Gaussian noise, and color jittering. Furthermore, we employ a weighted random sampling strategy to address the class imbalance between positive and negative samples. 

Upon completion of training, we perform a line search on the validation set to determine the optimal threshold for the radius $R$. During the inference phase, the center point $o$ and the radius $R$ are utilized to ascertain whether a given sample falls within the hypersphere.

\begin{table*}[]
    \centering
    \renewcommand{\arraystretch}{1.3}
    \setlength{\tabcolsep}{1.4mm}
    \label{tab:performance_comparison_rounded}
    \begin{tabular}{l l c c c c c}
        \hline\hline
        \multirow{2}{*}{Datasets} & \multirow{2}{*}{Methods} & SD1.5-Db & SD1.5-LoRA & SD2.1-Db & SD2.1-LoRA & Kandinsky \\
        \cline{3-7} 
        & & AUC / T@$10^{-2}$F & AUC / T@$10^{-2}$F & AUC / T@$10^{-2}$F & AUC / T@$10^{-2}$F & AUC / T@$10^{-2}$F \\
        \hline
        \multirow{5}{*}{WikiArt} 
        & RoSteALS      & 0.578 / 0.010  & 0.589 / 0.013   & 0.571 / 0.008  & 0.576 / 0.014  & 0.583 / 0.012  \\
        & DIAGNOSIS     & 0.901 / 0.256  & 0.767 / 0.153  & 0.831 / 0.139  & 0.840 / 0.231  & 0.802 / 0.120    \\
        & SIREN         & 0.721 / 0.196  & 0.580 / 0.016   & 0.809 / 0.237  & 0.564 / 0.031   & 0.699 / 0.067   \\
        & ArtistAuditor & 0.973 / 0.626  & 0.970 / 0.601  & 0.972 / 0.632  & 0.969 / 0.652  & 0.973 / 0.681  \\
        & Ours          & \textbf{0.993 / 0.928} & \textbf{0.994 / 0.938} & \textbf{0.994 / 0.925} & \textbf{0.998 / 0.953} & \textbf{0.998 / 0.949}  \\
        \hline
        \multirow{5}{*}{ArtBench} 
        & RoSteALS      & 0.575 / 0.018  & 0.574 / 0.012  & 0.571 / 0.015   & 0.565 / 0.014   & 0.554 / 0.012  \\
        & DIAGNOSIS     & 0.785 / 0.155 & 0.743 / 0.138  & 0.785 / 0.157  & 0.744 / 0.134 & 0.864 / 0.193 \\
        & SIREN         & 0.613 / 0.019  & 0.621 / 0.010  & 0.712 / 0.129  & 0.612 / 0.012  & 0.688 / 0.063  \\
        & ArtistAuditor & 0.918 / 0.470  & 0.916 / 0.506  & 0.921 / 0.576  & 0.912 / 0.505  & 0.942 / 0.634  \\
        & Ours          & \textbf{0.990 / 0.901} & \textbf{0.993 / 0.922} & \textbf{0.992 / 0.882} & \textbf{0.989 / 0.862} & \textbf{0.995 / 0.930} \\
        \hline\hline 
    \end{tabular}
    \caption{Performance comparison across five fine-tuning settings on the WikiArt and ArtBench datasets, where Db represents Dreambooth and T@$10^{-2}$F represents TPR@FPR$=10^{-2}$. The best result under each metric is marked with \textbf{bold}.}
    \label{main_result}
\end{table*}

\section{Experiment}
\subsection{Experimental Setting}
\textbf{Datasets and Models. }Our experiments are conducted on subsets of two large-scale art datasets: WikiArt \cite{tan2018improved} and ArtBench \cite{liao2022artbench}. To emulate unauthorized usage of artworks, we fine-tune a set of three surrogate text-to-image models with DreamBooth and LoRA, including Stable Diffusion v1.5 \cite{rombach2022high}, Stable Diffusion v2.1, and Kandinsky \cite{razzhigaev2023kandinsky}. Both Dreambooth and LoRA are applied to Stable Diffusion, while Kandinsky is fine-tuned exclusively with LoRA for memory efficiency.

\textbf{Implementation Details. }All experiments are conducted on an NVIDIA RTX 3090 GPU. Before training, we generate 1 to 3 augmented images for each positive sample. Negative samples are sourced from the WikiArt dataset. The verifier is trained with the AdamW optimizer (learning rate $5\times10^{-4}$) with key hyperparameters: $\lambda_{pos}=1.0$, $\lambda_{neg}=1.0$, $\beta=0.3$, and $m=1.0$. The test set is balanced, comprising 1,000 mimicked images and 1,000 unrelated style images. All attack models are fine-tuned using default parameters from their original implementations.

\textbf{Evaluation Metrics. }We use two primary metrics: the area under the curve (AUC) and the True Positive Rate at a $10^{-2}$ False Positive Rate (TPR@FPR$=10^{-2}$), where AUC quantifies the overall discriminative ability across all thresholds and TPR@FPR=$10^{-2}$ measures the verification sensitivity under strict constraints.

\textbf{Baseline. }We mainly compare our StyleSentinel with four state-of-the-art methods, including two watermarking-based methods SIREN \cite{li2025towards} and RoSteALS \cite{bui2023rosteals}, a backdoor-based method DIAGNOSIS \cite{wang2023diagnosis}, and a verification method ArtistAuditor \cite{du2025artistauditor}. To ensure a fair comparison in a unified way, we convert their native metrics to AUC and TPR@FPR$=10^{-2}$ inspired by \cite{li2025towards}.

\subsection{Performance Evaluation}

As demonstrated in Table \ref{main_result}, our StyleSentinel consistently and significantly outperforms four baselines across all tested settings. On the WikiArt dataset, our approach achieves the highest AUC scores between 0.993 and 0.998 while maintaining the highest on TPR@FPR$=10^{-2}$ between 0.925 and 0.953. These results demonstrate the remarkable robustness of StyleSentinel, validating the efficacy of using a stylistic fingerprint for copyright verification.
In contrast, the baselines either fail to effectively detect the unauthorized data usage or have very fluctuating performance across different settings. The suboptimal baseline ArtistAuditor achieves high AUC scores, but its performance on TPR@FPR$=10^{-2}$ represents a significant gap of over 25 percent compared to our method. The remaining baselines perform even worse. The backdoor-based method, DIAGNOSIS, shows unstable performance across different fine-tuning techniques, while the watermarking methods, SIREN and RoSteALS, are rendered almost entirely ineffective. Their extremely low AUC scores and TPR values confirm that their embedded signals are too fragile to survive the fine-tuning process and to be detected in one-sample scenarios.
These results confirm that intrinsic stylistic fingerprints provide more effective and robust verification evidence than the extrinsic signals used by watermark-based or backdoor-based methods.

\begin{figure}[t]
\includegraphics[width=\linewidth]{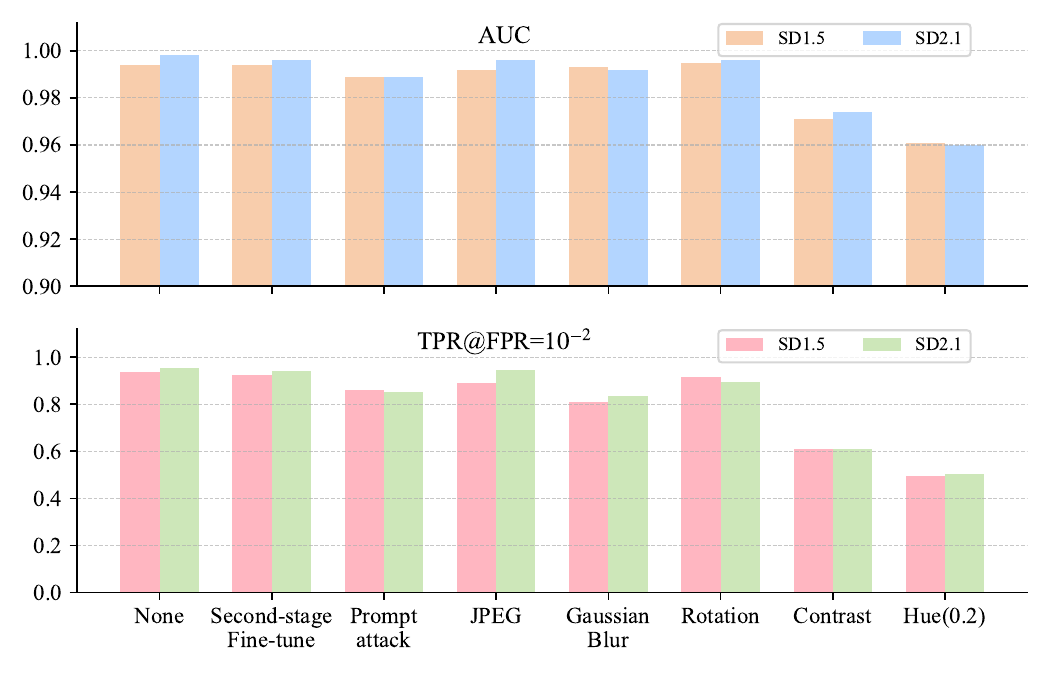}
\caption{Robustness of StyleSentinel against common image transformations and adaptive attacks. We evaluate the performance on Stable Diffusion v1.5 and v2.1.}
\label{robust}
\end{figure}

\begin{table}[h]
\centering
\renewcommand{\arraystretch}{1.3}
\setlength{\tabcolsep}{1.2mm}
\begin{tabular}{l cc cc}
\hline\hline
\multirow{2}{*}{Configuration} & \multicolumn{2}{c}{SD1.5} & \multicolumn{2}{c}{SD2.1} \\
\cline{2-5} 
& AUC & T@$10^{-2}$F & AUC & T@$10^{-2}$F \\
\hline
\textbf{Our Full Method} & \textbf{0.994} & \textbf{0.938} & \textbf{0.998} & \textbf{0.953} \\
\hline
w/o Aug. & 0.975 & 0.659 & 0.978 & 0.707 \\
w/ Trad. Aug. & 0.988 & 0.791 & 0.986 & 0.797 \\
\hline
Concatenation & 0.987 & 0.803 & 0.978 & 0.741 \\
\hline
BCE Loss & 0.992 & 0.838 & 0.989 & 0.851 \\
\hline\hline
\end{tabular}
\caption{Ablation study on method components.}
\label{ablation1}
\end{table}

Notably, when migrating to the lower-resolution ($256\times256$) ArtBench dataset, the performance of most baseline methods declines markedly due to inferior fine-tuning results. Although our method is also slightly affected, it consistently maintains the highest performance, demonstrating superior robustness to variations in image quality. Furthermore, StyleSentinel exhibits uniform verification capabilities across different fine-tuning settings on both datasets. This resilience stems from the ability of StyleSentinel to extract overall and intrinsic stylistic features, which are robust and transferable.
\begin{table*}[t]
    \centering
    \renewcommand{\arraystretch}{1.3}
    \setlength{\tabcolsep}{1.8mm}
    \begin{tabular}{l cc cc | ccc} 
        \hline\hline
        \multirow{3}{*}{Dataset} & \multicolumn{4}{c|}{\textbf{Completely Disjoint}} & \multicolumn{3}{c}{\textbf{Partially Overlapping}} \\ 
        \cline{2-8}
        & \multicolumn{2}{c}{SD1.5-Db} & \multicolumn{2}{c|}{\text{SD2.1-Db}} & \multicolumn{1}{c}{SD1.5-LoRA} & \multicolumn{1}{c}{SD2.1-LoRA} & \multicolumn{1}{c}{Kandinsky} \\ 
        \cline{2-8}
        & AUC & T@$10^{-2}$F & AUC & T@$10^{-2}$F & AUC / T@$10^{-2}$F & AUC / T@$10^{-2}$F & AUC / T@$10^{-2}$F \\
        \hline
        WikiArt  & 0.991 & 0.810 & 0.990 & 0.793 & 0.997 / 0.942 & 0.995 / 0.938 & 0.996 / 0.900 \\
        ArtBench & 0.992 & 0.846 & 0.988 & 0.857 & 0.994 / 0.885 & 0.995 / 0.895 & 0.994 / 0.886 \\
        \hline\hline
    \end{tabular}
    \caption{Generalization study of StyleSentinel. We test against models fine-tuned with completely disjoint dataset for DreamBooth and partially overlapping dataset for LoRA. The excellent performance shows strong generalization of StyleSentinel.}
    \label{generalization}
\end{table*}

\subsection{Robustness Study}

We evaluate the robustness of StyleSentinel against two primary categories of adversarial manipulations: common image transformations encountered during online distribution and sophisticated adaptive attacks. The former category comprises five distinct transformations: rotation, JPEG compression (50\% quality), Gaussian blurring (3x3 kernel), color jittering (hue factor of 0.2) and contrast adjustment (factor of 2.0). The adaptive attack strategies included two-stage fine-tuning on generated images and black-box prompt attacks employing different textual descriptions.

As shown in Figure \ref{robust}, StyleSentinel demonstrates remarkable robustness against common transformations and adaptive attacks. It exhibits high resilience to geometric manipulations and compression artifacts. 
Although alterations in the color space influence style features and cause a decline in TPR@FPR$=10^{-2}$, the consistently high AUC demonstrates retained robustness in verification performance.
Moreover, the method effectively withstands adaptive strategies such as second-stage fine-tuning and maintains its efficacy against prompt attacks. This resilience against both superficial transformations and adaptive attacks suggests that the learned fingerprint captures the deep feature of an artist's style, rather than just low-level textures.

\begin{figure}[t]
\includegraphics[width=0.99\linewidth]{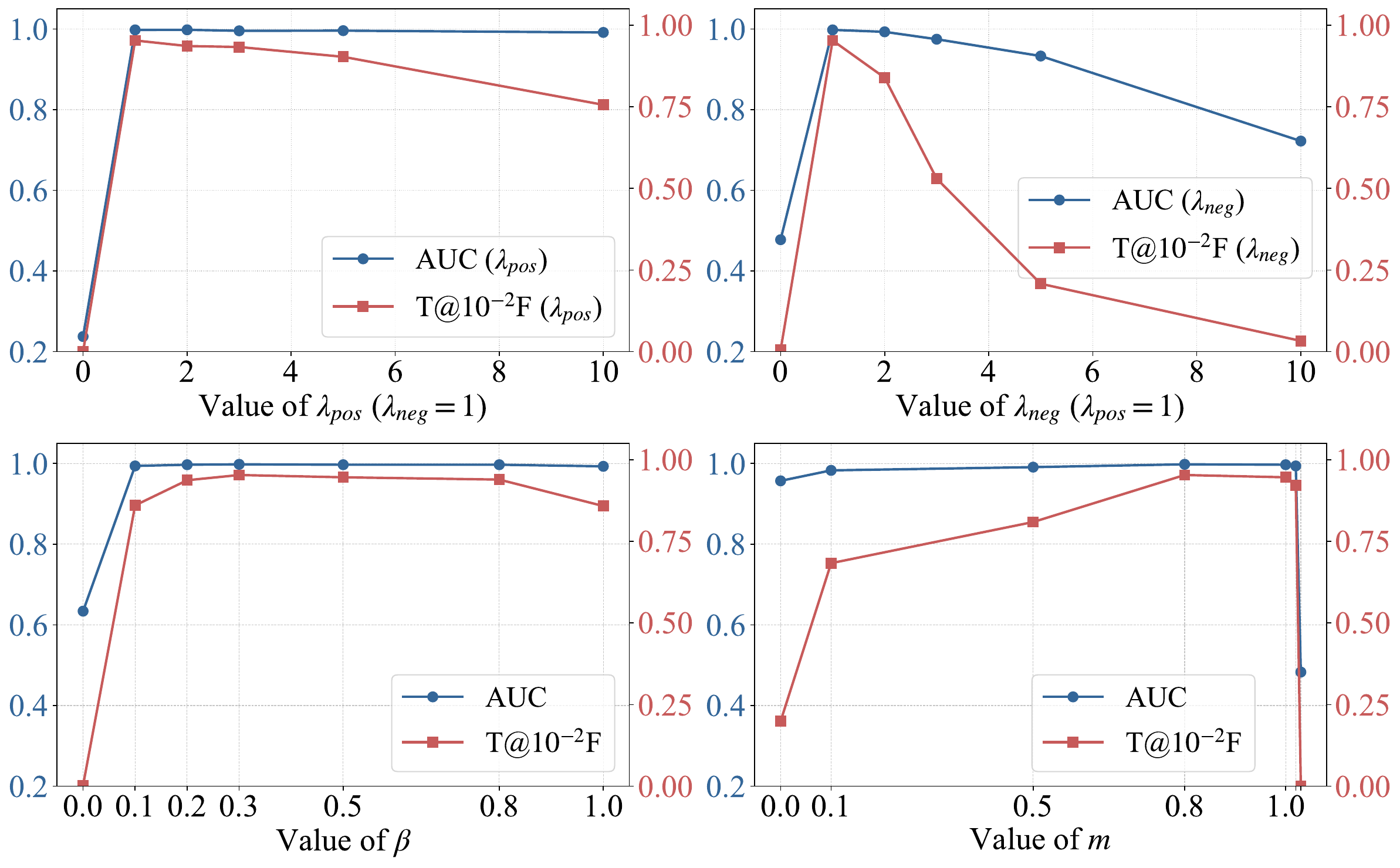}
\caption{Ablation study on hyperparameters.}
\label{ablation2}
\end{figure}

\subsection{Generalization Study}
We evaluate the generalization of StyleSentinel when the artworks used to train the verifier differ from those used to fine-tune the suspicious model. Specifically, we use a partially overlapping dataset for the suspicious model fine-tuned with LoRA, and a completely disjoint dataset for the one fine-tuned with Dreambooth.
As demonstrated in Table \ref{generalization}, our method shows remarkable generalization, consistently achieving AUC scores around 0.99 and high TPR@FPR$=10^{-2}$ values ranging from 0.793 to 0.942 in both the WikiArt and ArtBench datasets. 
The consistent performance across all settings confirms that our approach learns a transferable stylistic fingerprint, representing the intrinsic rules of the artist's style.

\subsection{Ablation Study}

\textbf{Component Ablation. }As shown in Table \ref{ablation1}, we systematically validate the contribution of each component. 
First, our semantic self-reconstruction style augmentation demonstrates a significant performance improvement compared to both a no-augmentation baseline and traditional augmentation methods, confirming its effectiveness for subsequent style feature learning. 
Furthermore, replacing the attentional fusion module with simple feature concatenation leads to a performance decline, which highlights the necessity of adaptively fusing features from different hierarchical levels.
Finally, substituting the hypersphere-based verifier with a standard BCE-loss binary classifier results in degraded performance,
\begin{table}[!t]
\centering
\renewcommand{\arraystretch}{1.3} 
\setlength{\tabcolsep}{3pt}   
\begin{tabular}{l cc cc}
    \hline\hline
    \multirow{2}{*}{Platform} & \multicolumn{2}{c}{WikiArt} & \multicolumn{2}{c}{ArtBench} \\
    \cline{2-5} 
    & AUC & T@$10^{-2}$F & AUC & T@$10^{-2}$F \\
    \hline
    Shakker  & 0.998 & 0.944 & 0.994 & 0.933 \\
    LibLibAI & 0.994 & 0.928 & 0.993 & 0.922 \\
    \hline\hline
\end{tabular}
\caption{Real-world performance of StyleSentinel. We mainly test the effect on two common online platforms.}
\label{real_world}
\end{table}
validating that a one-class learning approach is more suitable for stylistic fingerprint verification.

\textbf{Hyperparameter Sensitivity. }As shown in Figure \ref{ablation2}, we reveal several key sensitivities of the hyperparameters. 
We found that balancing the loss weights at $\lambda_{pos}=\lambda_{neg}=1.0$ is critical, as any imbalance leads to a performance degradation. Similarly, the repulsion intensity $\beta$ has an optimal value at $0.3$, which best separates the classes without creating an overly rigid boundary. The soft margin $m$ also shows a clear trade-off with performance peaking between 0.8 and 1.0, but excessively high values ($\textgreater 1.0$) introduce training instability, resulting in significant performance degradation.

\subsection{Real-World Performance}
We demonstrate the effectiveness of StyleSentinel in two real-world online fine-tuning service, Shakker\footnote{https://www.shakker.ai} and LibLibAI\footnote{https://www.liblib.art}. The two platforms allow users to fine-tune a model with their own uploaded images and provide an API endpoint for generating mimicked images. We conducted tests on WikiArt and ArtBench using the default models and fine-tuning methods offered by this service. As shown in Table \ref{real_world}, our method maintained an AUC of more than 0.99 and a TPR of more than 0.90 in an FPR of $10^{-2}$, proving its reliability in practical applications.

\section{Conclusion}
In this paper, we introduced StyleSentinel, a novel approach for artistic copyright verification that operates by learning an intrinsic stylistic fingerprint directly from the artworks. It requires no preprocessing and enables protection for artworks already in circulation online.
Our approach employs a semantic self-reconstruction process to overcome data sparsity and uses a multi-layer attention style extractor to encode the stylistic fingerprint. Moreover, it formulates the verification task as a robust one-class learning problem with a hypersphere-based style verifier. Extensive experiments demonstrated that StyleSentinel outperforms state-of-the-art baselines in challenging one-sample scenarios. The effectiveness of our method was further validated on two real-world online platforms. We believe StyleSentinel offers a reliable verification for artistic copyright protection.


\bibliography{anonymous}

\end{document}